\documentclass[11pt,a4paper]{article}
\usepackage[hyperref]{naaclhlt2019}
\def\aclpaperid{1123}
\usepackage{times}
\usepackage{latexsym}

\usepackage{balance}
\usepackage{amssymb}
\usepackage{amsmath}
\usepackage{bm}
\usepackage{color}
\usepackage{multirow}
\usepackage{subfig}
\usepackage{arydshln}
\usepackage{amsfonts}
\usepackage{graphicx} 
\usepackage{graphics}
\usepackage{standalone}
\usepackage{caption}

\aclfinalcopy 

\usepackage{color}
\definecolor{zptu}{RGB}{18, 141, 21}

\title{Convolutional Self-Attention Networks}

\author{Baosong Yang$^\dagger$~~~~~Longyue Wang$^\ddagger$~~~~~Derek F. Wong$^{\dagger}$~~~~~Lidia S. Chao$^\dagger$~~~~~Zhaopeng Tu$^\ddagger$\thanks{~Zhaopeng Tu is the corresponding author of the paper. This work was conducted when Baosong Yang was interning at Tencent AI Lab.}\\
  $^\dagger$NLP$^2$CT Lab, Department of Computer and Information Science,
  University of Macau\\
  {\tt nlp2ct.baosong@gmail.com, \{derekfw,lidiasc\}@umac.mo} \\
  $^\ddagger$Tencent AI Lab\\
  {\tt \{vinnylywang,zptu\}@tencent.com}}

\begin{document}
\maketitle
\begin{abstract}
Self-attention networks (SANs) have drawn increasing interest due to their high parallelization in computation and flexibility in modeling dependencies. SANs can be further enhanced with multi-head attention by allowing the model to attend to information from different representation subspaces.
In this work, we propose novel {\em convolutional self-attention networks}, which offer SANs the abilities to 1) strengthen dependencies among neighboring elements, and 2) model the interaction between features extracted by multiple attention heads.
Experimental results of machine translation on different language pairs and model settings show that our approach outperforms both the strong Transformer baseline and other existing models on enhancing the locality of SANs. Comparing with prior studies, the proposed model is parameter free in terms of introducing no more parameters.

\end{abstract}


\section{Introduction}

Self-attention networks (SANs) \cite{parikh2016decomposable,lin2017structured} have shown promising empirical results in various natural language processing (NLP) tasks, such as machine translation~\cite{Vaswani:2017:NIPS}, natural language inference~\cite{Shen:2018:AAAI}, and acoustic modeling~\cite{sperber2018self}. 
One appealing strength of SANs lies in their ability to capture dependencies regardless of distance by explicitly attending to all the elements.
In addition, the performance of SANs can be improved by multi-head attention~\cite{Vaswani:2017:NIPS}, which projects the input sequence into multiple subspaces and applies attention to the representation in each subspace.

Despite their success, SANs have 
two major limitations. First, the model fully take into account all the elements, which  disperses the attention distribution and thus overlooks the relation of neighboring elements and phrasal patterns~\cite{Yang:2018:EMNLP,wu2018phrase,guo2019gaussian}. 
Second, multi-head attention 
 extracts distinct linguistic properties from each subspace in a parallel fashion~\cite{Raganato:2018:EMNLPWorkshop}, which fails to exploit useful interactions across different heads. Recent work shows that better features can be learned if different sets of representations are present at feature learning time~\cite{ngiam2011multimodal,lin2013network}. 

 To this end, we propose novel {\em convolutional self-attention networks} (\textsc{Csan}s), which 
 model locality for self-attention model and interactions between features learned by different attention heads in an unified framework.
Specifically, in order to pay more attention to a local part of the input sequence, we restrict the attention scope to a window of neighboring elements. The localness is therefore enhanced via a parameter-free 1-dimensional convolution. 
Moreover, we extend the convolution to a 2-dimensional area with the axis of attention head. Thus, the proposed model allows each head to interact local features with its adjacent subspaces at attention time. We expect that the interaction across different subspaces can further improve the performance of SANs.

\begin{figure*}[t]
\begin{center}
\subfloat[Vanilla SANs]{\includegraphics[width=0.6\columnwidth]{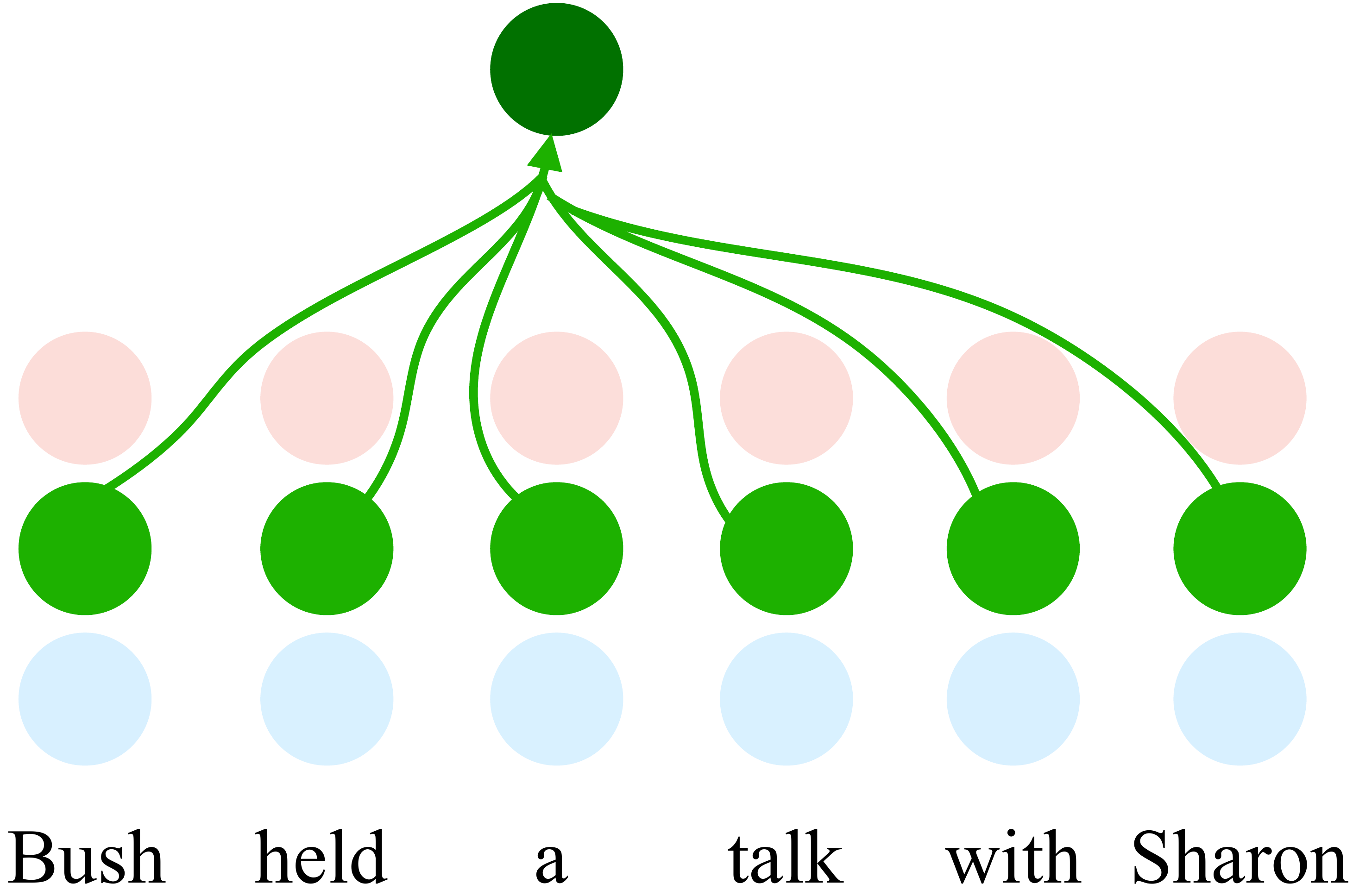}}\hspace{0.1\columnwidth}
\subfloat[1D-Convolutional SANs]{\includegraphics[width=0.6\columnwidth]{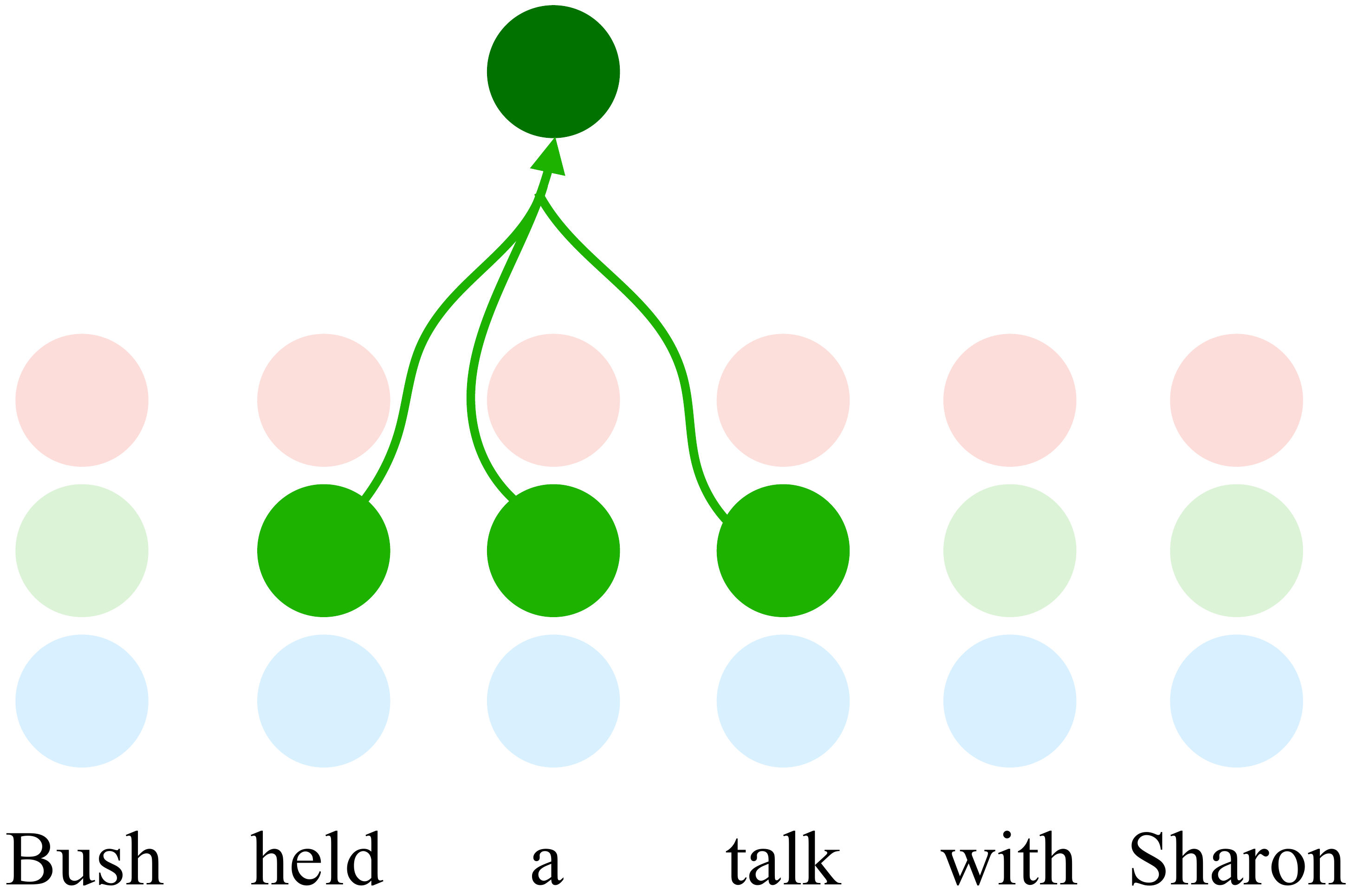}}\hspace{0.1\columnwidth}
\subfloat[2D-Convolutional SANs]{\includegraphics[width=0.6\columnwidth]{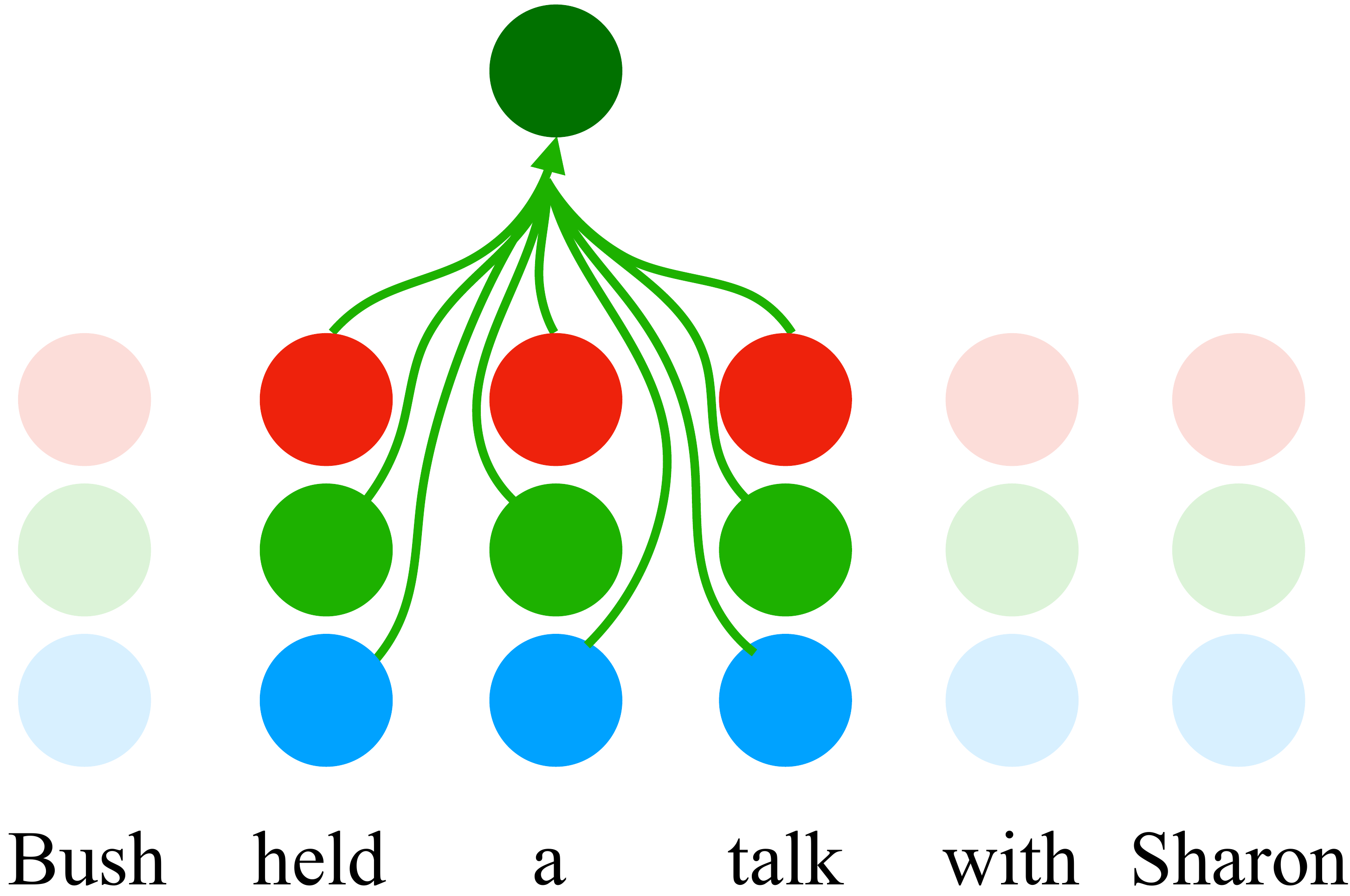}}
\caption{Illustration of (a) vanilla SANs; (b) 1-dimensional convolution with the window size being $3$; and (c) 2-dimensional convolution with the area being $3 \times 3$. Different colors represent different subspaces modeled by multi-head attention, and transparent colors denote masked tokens that are invisible to SANs.}

\label{fig:csan}
\end{center}
\end{figure*}

We evaluate the effectiveness of the proposed model on three widely-used translation tasks: WMT14 English-to-German, WMT17 Chinese-to-English, and WAT17 Japanese-to-English. Experimental results demonstrate that our approach consistently improves performance over the strong \textsc{Transformer} model~\cite{Vaswani:2017:NIPS} across language pairs. 
Comparing with previous work on modeling locality for SANs~\citep[e.g.][]{shaw2018self,Yang:2018:EMNLP,sperber2018self}, our model boosts performance on both translation quality and training efficiency.


\section{Multi-Head Self-Attention Networks}

SANs produce representations by applying attention to each pair of tokens from the input sequence, regardless of their distance.~\newcite{Vaswani:2017:NIPS} found it is beneficial to capture different contextual features with multiple individual attention functions. Given an input sequence $\mathbf{X} = \{\mathbf{x}_1,\dots,\mathbf{x}_I\} \in \mathbb{R}^{I \times d}$, the model first transforms it into queries $\mathbf{Q}$, keys $\mathbf{K}$, and values $\mathbf{V}$:
\begin{equation}
    \mathbf{Q}, \mathbf{K}, \mathbf{V} = \mathbf{X} \mathbf{W}_Q, \mathbf{X} \mathbf{W}_K, \mathbf{X} \mathbf{W}_V \in \mathbb{R}^{I \times d}
\end{equation}
where $\{\mathbf{W}_Q, \mathbf{W}_K, \mathbf{W}_V\} \in \mathbb{R}^{d \times d}$ are trainable parameters and $d$ indicates the hidden size. The three types of representations are split into $H$ different subspaces, e.g., $[\mathbf{Q}^1, \dots, \mathbf{Q}^H] = \mathbf{Q}$ with $\mathbf{Q}^h \in \mathbb{R}^{I \times \frac{d}{H}}$. In each subspace $h$, the element $\mathbf{o}^h_i$ in the output sequence $\mathbf{O}^h = \{\mathbf{o}^h_1, \dots, \mathbf{o}^h_I\}$ is calculated by
\begin{align}
\label{eq:att}
    \mathbf{o}^h_i = \textsc{Att}(\mathbf{q}^h_i, \mathbf{K}^h) \mathbf{V}^h &&& \in \mathbb{R}^{\frac{d}{H}}
\end{align}
where $\textsc{Att}(\cdot)$ is an attention model~\cite{bahdanau2015neural,Vaswani:2017:NIPS} that retrieves the keys $\mathbf{K}^h$ with the query $\mathbf{q}^h_i$. The final output representation $\mathbf{O}$ is the concatenation of outputs generated by multiple attention models: 
\begin{align}
    \mathbf{O} = [\mathbf{O}^1, \dots, \mathbf{O}^H] && \in \mathbb{R}^{I \times d}
\end{align}

\section{Approach}

As shown in Figure~\ref{fig:csan}(a), the vanilla SANs use the query $\mathbf{q}_i^h$ to compute a categorical distribution over all elements from  $\mathbf{K}^h$~(Equation~\ref{eq:att}). It may inherit the attention to neighboring information~\cite{Yu:2018:ICLR,Yang:2018:EMNLP,guo2019gaussian}. In this work, we propose to model locality for SANs by restricting the model to attend to a local region via convolution operations (1D-\textsc{CSan}s, Figure~\ref{fig:csan}(b)). Accordingly, it provides distance-aware information (e.g. phrasal patterns), which is complementary to the distance-agnostic dependencies modeled by the standard SANs (Section~\ref{sec-convolution}). 

Moreover, 
the calculation of output $\mathbf{o}^h$ are restricted to the a single individual subspace, overlooking the richness of contexts and the dependencies among groups of features, which have proven beneficial to the feature learning~\cite{ngiam2011multimodal,wu2018group}. We thus propose to convolute the items in adjacent heads (2D-\textsc{CSan}s, Figure~\ref{fig:csan}(c)). The proposed model is expected to improve performance through interacting linguistic properties across heads (Section~\ref{sec-interaction}).

\subsection{Locality Modeling via 1D Convolution}
\label{sec-convolution}

For each query $\mathbf{q}^h_i$, we restrict its attention region (e.g., ${\bf K}^h = \{\mathbf{k}^h_1, \dots, \mathbf{k}^h_i, \dots, \mathbf{k}^h_I \}$) to a local scope with a fixed size $M+1$ ($M \le I$) centered at the position $i$:
\begin{eqnarray}
    \widehat{\mathbf{K}}^h &=& \{\mathbf{k}^h_{i-\frac{M}{2}}, \dots, \mathbf{k}^h_i, \dots, \mathbf{k}^h_{i+\frac{M}{2}} \} \label{eqn-1D_key} \\
    \widehat{\mathbf{V}}^h &=& \{\mathbf{v}^h_{i-\frac{M}{2}}, \dots, \mathbf{v}^h_i, \dots, \mathbf{v}^h_{i+\frac{M}{2}} \} \label{eqn-1D_value}
\end{eqnarray}
Accordingly, the calculation of corresponding output in Equation~(\ref{eq:att}) is modified as:
\begin{eqnarray}
    \mathbf{o}^h_i = \textsc{Att}(\mathbf{q}^h_i, \widehat{\mathbf{K}}^h) \widehat{\mathbf{V}}^h
    \label{eq:conc}
\end{eqnarray}
As seen, SANs are only allowed to attend to the neighboring tokens (e.g., $\widehat{\mathbf{K}}^h$, $\widehat{\mathbf{V}}^h$), instead of all the tokens in the sequence (e.g., $\mathbf{K}^h$, $\mathbf{V}^h$).

The SAN-based models are generally implemented as multiple layers, in which higher layers tend to learn semantic information while lower layers capture surface and lexical information~\cite{Peters:2018:NAACL,Raganato:2018:EMNLPWorkshop}.  Therefore, we merely apply locality modeling to the lower layers, which same to the configuration in~\newcite{Yu:2018:ICLR} and~\newcite{Yang:2018:EMNLP}. In this way, the representations are learned in a hierarchical fashion~\cite{yang2017towards}. That is, the distance-aware and local information extracted by the lower SAN layers, is expected to complement distance-agnostic and global information captured by the higher SAN layers.

\subsection{Attention Interaction via 2D Convolution}
\label{sec-interaction}
Mutli-head mechanism allows different heads to capture distinct linguistic properties~\cite{Raganato:2018:EMNLPWorkshop,Li:2018:EMNLP}, especially in diverse local contexts~\cite{sperber2018self,Yang:2018:EMNLP}. We hypothesis that  exploiting  local properties across heads is able to further improve the performance of SANs. 
To this end, we expand the 1-dimensional window to a 2-dimensional area with the new dimension being the index of attention head. Suppose that the area size is $(N+1) \times (M+1)$ ($N \le H$), the keys and values in the area are:
\begin{eqnarray}
    \widetilde{\mathbf{K}}^h &=&  \bigcup [\widehat{\mathbf{K}}^{h-\frac{N}{2}}, \dots, \widehat{\mathbf{K}}^{h}, \dots,  \widehat{\mathbf{K}}^{h+\frac{N}{2}}] \\
    \widetilde{\mathbf{V}}^h &=&  \bigcup [\widehat{\mathbf{V}}^{h-\frac{N}{2}}, \dots, \widehat{\mathbf{V}}^{h}, \dots, \widehat{\mathbf{V}}^{h+\frac{N}{2}}] 
\end{eqnarray}
where $\widehat{\mathbf{K}}^h, \widehat{\mathbf{V}}^h$ are elements in the $h$-th subspace, which are calculated by Equations~\ref{eqn-1D_key} and~\ref{eqn-1D_value} respectively. The union operation $\bigcup$ means combining the keys and values in different subspaces. 
The corresponding output is calculated as:
\begin{eqnarray}
    \mathbf{o}^h_i = \textsc{Att}(\mathbf{q}^h_i, \widetilde{\mathbf{K}}^h) \widetilde{\mathbf{V}}^h
\end{eqnarray}
The 2D convolution allows SANs to build relevance between elements across adjacent heads, thus flexibly extract local features from different subspaces rather than merely from an unique head. 

The vanilla SAN models linearly aggregate features from different heads, and this procedure limits the extent of abstraction~\cite{fukui2016multimodal,Li:2019:NAACL}. Multiple sets of representations presented at feature learning time can further improve the expressivity of the learned features~\cite{ngiam2011multimodal,wu2018group}.

\begin{table*}[t]
\begin{center}
\begin{tabular}{l| r c |cc}
    {\bf Model}  & {\bf Parameter} & {\bf Speed}  & {\bf BLEU} & \bf $\triangle$ \\
    \hline \hline
    \textsc{Transformer-Base} \cite{Vaswani:2017:NIPS} & 88.0M & 1.28 & 27.31 & - \\ 
    \hline
    ~~+ \textsc{Bi\_Direct} \cite{Shen:2018:AAAI}  & +0.0M & -0.00 & 27.58 & +0.27 \\  
    ~~+ \textsc{Rel\_Pos} \cite{shaw2018self}  & +0.1M & -0.11  & 27.63 & +0.32   \\ 
    ~~+ \textsc{Neighbor} \cite{sperber2018self}  & +0.4M & -0.06   & 27.60  & +0.29   \\ 
    ~~+ \textsc{Local\_Hard} \cite{luong2015effective}  & +0.4M & -0.06   & 27.73 & +0.42 \\  
    ~~+ \textsc{Local\_Soft} \cite{Yang:2018:EMNLP}  & +0.8M & -0.09  & 27.81 & +0.50 \\ 
    ~~+ \textsc{Block} \cite{Shen:2018:ICLR}  & +6.0M & -0.33   & 27.59 & +0.28 \\  
    ~~+ \textsc{Cnn}s \cite{Yu:2018:ICLR}  & +42.6M & -0.54 & 27.70 & +0.39   \\ 
    \hline
    ~~+ 1D-\textsc{CSan}s         & +0.0M & -0.00  & \bf 27.86 & +0.55    \\ 
    ~~+ 2D-\textsc{CSan}s       & +0.0M & -0.06   & \textbf{28.18}  & +0.87    \\ 
\end{tabular}
\caption{Comparing with the existing approaches on WMT14 En$\Rightarrow$De translation task. For a fair comparison, we re-implemented the existing locality approaches under the same framework. ``Parameter" denotes the number of model parameters (M = million) and ``Speed'' denotes the training speed (steps/second). ``$\bigtriangleup$'' column denotes performance improvements over the Transformer baseline.}
\label{table:comparison}
\end{center}
\end{table*}

\section{Related Work}
\label{sec:related}

\paragraph{Self-Attention Networks} Recent studies have shown that \textsc{San}s can be further improved by capturing complementary information. For example,~\newcite{Chen:2018:ACL} and~\newcite{Hao:2019:NAACL} complemented \textsc{San}s with recurrence modeling, while~\newcite{Yang:2019:AAAI} modeled contextual information for \textsc{San}s.

Concerning modeling locality for \textsc{San}s,~\newcite{Yu:2018:ICLR} injected several CNN layers \cite{kim2014convolutional} to fuse local information, the output of which is fed to the subsequent SAN layer. Several researches proposed to revise the attention distribution with a parametric localness bias, and succeed on machine translation~\cite{Yang:2018:EMNLP} and natural language inference~\cite{guo2019gaussian}.
While both models introduce additional parameters, our approach is a more lightweight solution without introducing any new parameters. 
Closely related to this work,~\newcite{Shen:2018:AAAI} applied a positional mask to encode temporal order, which only allows SANs to attend to the previous or following tokens in the sequence. In contrast, we employ a positional mask (i.e. the tokens outside the local window is masked as $0$) to encode the distance-aware local information.

In the context of distance-aware SANs,~\newcite{shaw2018self} introduced relative position encoding to consider the relative distances between sequence elements. While they modeled locality from position embedding, we improve locality modeling from revising attention scope. 
To make a fair comparison, we re-implemented the above approaches under a same framework. 
Empirical results on machine translation tasks show 
the  superiority  of our approach 
in both translation quality and training efficiency.

\paragraph{Multi-Head Attention}
Multi-head attention mechanism~\cite{Vaswani:2017:NIPS} employs different attention heads to capture distinct features~\cite{Raganato:2018:EMNLPWorkshop}.
Along this direction,~\newcite{Shen:2018:AAAI} explicitly used multiple attention heads to model different dependencies of the same word pair, and~\newcite{Strubell:2018:EMNLP} employed different attention heads to capture different linguistic features.
~\newcite{Li:2018:EMNLP} introduced disagreement regularizations to encourage the diversity among attention heads.
Inspired by recent successes on fusing information across layers~\cite{Dou:2018:EMNLP,Dou:2019:AAAI},~\newcite{Li:2019:NAACL} proposed to aggregate information captured by different attention heads.
Based on these findings, we model interactions among attention heads to exploit the richness of local properties distributed in different heads.

\section{Experiments}

We conducted experiments with the Transformer model~\cite{Vaswani:2017:NIPS} on 
English$\Rightarrow$German (En$\Rightarrow$De), Chinese$\Rightarrow$English (Zh$\Rightarrow$En) and Japanese$\Rightarrow$English (Ja$\Rightarrow$En) translation tasks.  
For the En$\Rightarrow$De and Zh$\Rightarrow$En tasks, the models were trained on widely-used WMT14 and WMT17 corpora, consisting of around $4.5$ and $20.62$ million sentence pairs, respectively. 
Concerning~Ja$\Rightarrow$En, we followed~\newcite{morishita2017ntt} to use the first two sections of WAT17 corpus as the training data, which consists of 2M sentence pairs.  
To reduce the vocabulary size, all the data were tokenized and segmented into subword symbols using byte-pair encoding \cite{sennrich2015neural} with 32K merge operations. 
Following \newcite{shaw2018self}, we incorporated the proposed model into the encoder, which is a stack of 6 SAN layers.
Prior studies revealed that modeling locality in lower layers can achieve better performance~\cite{Shen:2018:ICLR,Yu:2018:ICLR,Yang:2018:EMNLP}, we applied our approach to the lowest three layers of the encoder. About configurations of NMT models, we used the \emph{Base} and \emph{Big} settings same as \newcite{Vaswani:2017:NIPS}, and all models were trained on 8 NVIDIA P40 GPUs with a batch of 4096 tokens.


\begin{figure}[t]
\begin{center}
\subfloat[1D-\textsc{CSan}s]{
\includegraphics[width=0.36\textwidth]{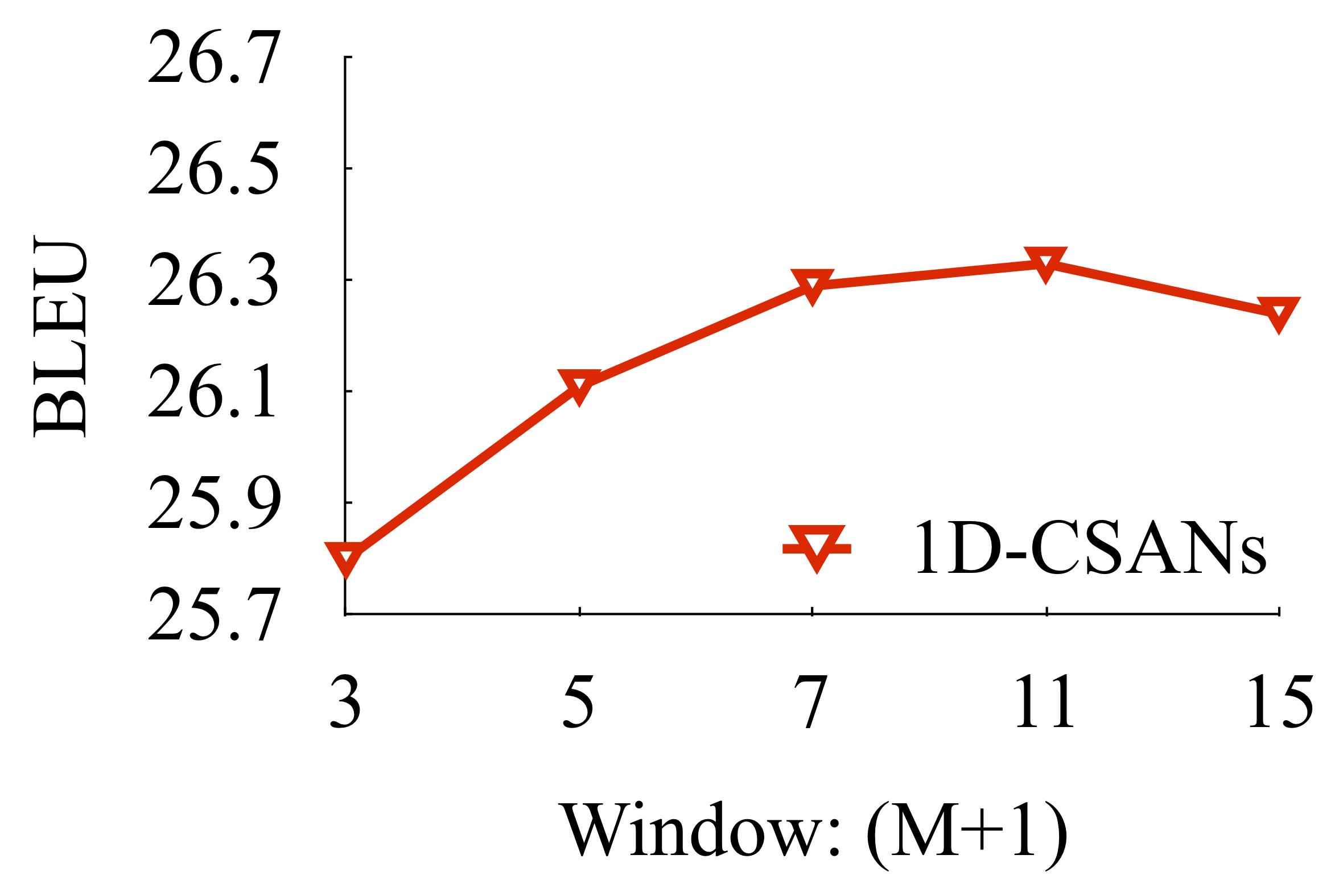}} \hspace{0.03\columnwidth}
\subfloat[2D-\textsc{CSan}s]{
\includegraphics[width=0.36\textwidth]{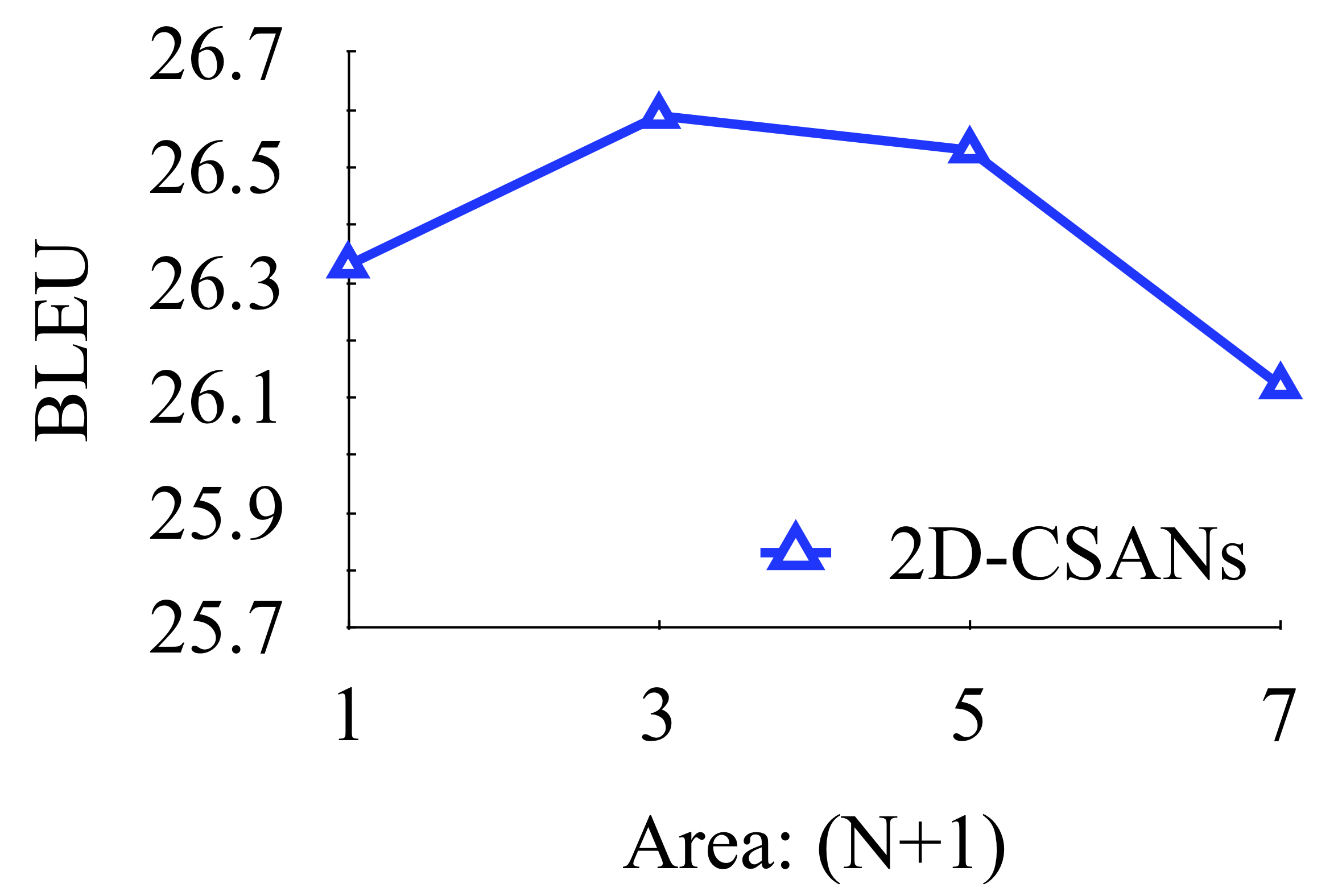}}
\caption{Effects of (a) window size on 1D-\textsc{CSan}s, and (b) attended head numbers on 2D-\textsc{CSan}s. For 2D-\textsc{CSan}s, the window size dimension is fixed to be 11.} 
\label{fig:size}
\end{center}
\end{figure}

\subsection{Effects of Window/Area Size}

\begin{table*}[t]
  \centering
  \begin{tabular}{l||rl|rl|rl}
     \multirow{2}{*}{\bf Model}  &  \multicolumn{2}{c}{\bf  WMT14 En$\Rightarrow$De}  & \multicolumn{2}{|c}{\bf WMT17 Zh$\Rightarrow$En}  &  \multicolumn{2}{|c}{\bf WAT17 Ja$\Rightarrow$En}\\ 
     \cline{2-7}
       & Speed  &   BLEU    &  Speed  &   BLEU  &  Speed  &   BLEU \\ 
    \hline \hline
       \textsc{Transformer-Base} & 1.28 &  27.31 & 1.21 & 24.13 & 1.33 & 28.10 \\ 
       ~~ + \textsc{CSan}s   &  1.22   & 28.18$^\Uparrow$ & 1.16 &  24.80$^\Uparrow$ & 1.28 & 28.50$^\uparrow$\\ 
    \hline
       \textsc{Transformer-Big}	 & 0.61 &   28.58  & 0.58 &  24.56 & 0.65 & 28.41 \\ 
       ~~ + \textsc{CSan}s  &  0.50 & 28.74 & 0.48 & 25.01$^\uparrow$  & 0.55 & 28.73$^\uparrow$ \\ 
  \end{tabular}
  \caption{Experimental results on WMT14 En$\Rightarrow$De, WMT17 Zh$\Rightarrow$En and WAT17 Ja$\Rightarrow$En test sets. ``Speed'' denotes the training speed (steps/second). ``$\uparrow/\Uparrow$'' indicates statistically significant difference from the vanilla self-attention counterpart ($ p < 0.05/0.01$), tested by bootstrap resampling~\cite{Koehn2004Statistical}.} 
  \label{tab:exist}
\end{table*}
We first investigated the effects of window size (1D-\textsc{CSan}s) and area size (2D-\textsc{CSan}s) on En$\Rightarrow$De validation set, as plotted in Figure~\ref{fig:size}. For 1D-\textsc{CSan}s, the local size with 11 is superior to other settings. This is consistent with \newcite{luong2015effective} 
who found that 10 is the best window size in their local attention experiments. 
Then, we fixed the number of neighboring tokens being 11 and varied the number of heads. As seen, by considering the features across heads (i.e. $>1$), 2D-\textsc{CSan}s further improve the translation quality. 
{However, when the number of heads in attention goes up, the translation quality inversely drops. One possible reason is that the model still has the flexibility of learning a different distribution for each head with few interactions, while a large amount of interactions assumes more heads make ``similar contributions'' \cite{wu2018group}.}




\subsection{Comparison to Related Work}
We re-implemented and compared several exiting works (Section~\ref{sec:related}) upon the same framework. 
Table \ref{table:comparison} lists the results on the En$\Rightarrow$De translation task.
As seen, all the models improve translation quality, reconfirming the necessity of modeling locality and distance information. Besides, our models outperform all the existing works, indicating the superiority of the proposed approaches. In particular, \textsc{CSan}s achieve better performance than \textsc{Cnn}s, revealing that extracting local features with dynamic weights 
(\textsc{CSan}s) is superior to assigning fixed parameters (\textsc{Cnn}s). 
Moreover, while most of the existing approaches (except for~\newcite{Shen:2018:AAAI}) introduce new parameters, our methods are parameter-free and thus only marginally affect training efficiency.

\subsection{Universality of The Proposed Model}
To validate the universality of our approach on MT tasks, we evaluated the proposed approach on different language pairs and model settings.
Table~\ref{tab:exist} lists the results on En$\Rightarrow$De, Zh$\Rightarrow$En and  Ja$\Rightarrow$En translation tasks. As seen, our model consistently improves translation performance across language pairs, which demonstrates the effectiveness and universality of the proposed approach. It is encouraging to see that  2D-Convolution with {\em base} setting  yields comparable performance with \textsc{Transformer-Big}.

\subsection{Accuracy of Phrase Translation}

\begin{figure}[t]
\begin{center}
\includegraphics[width=0.42\textwidth]{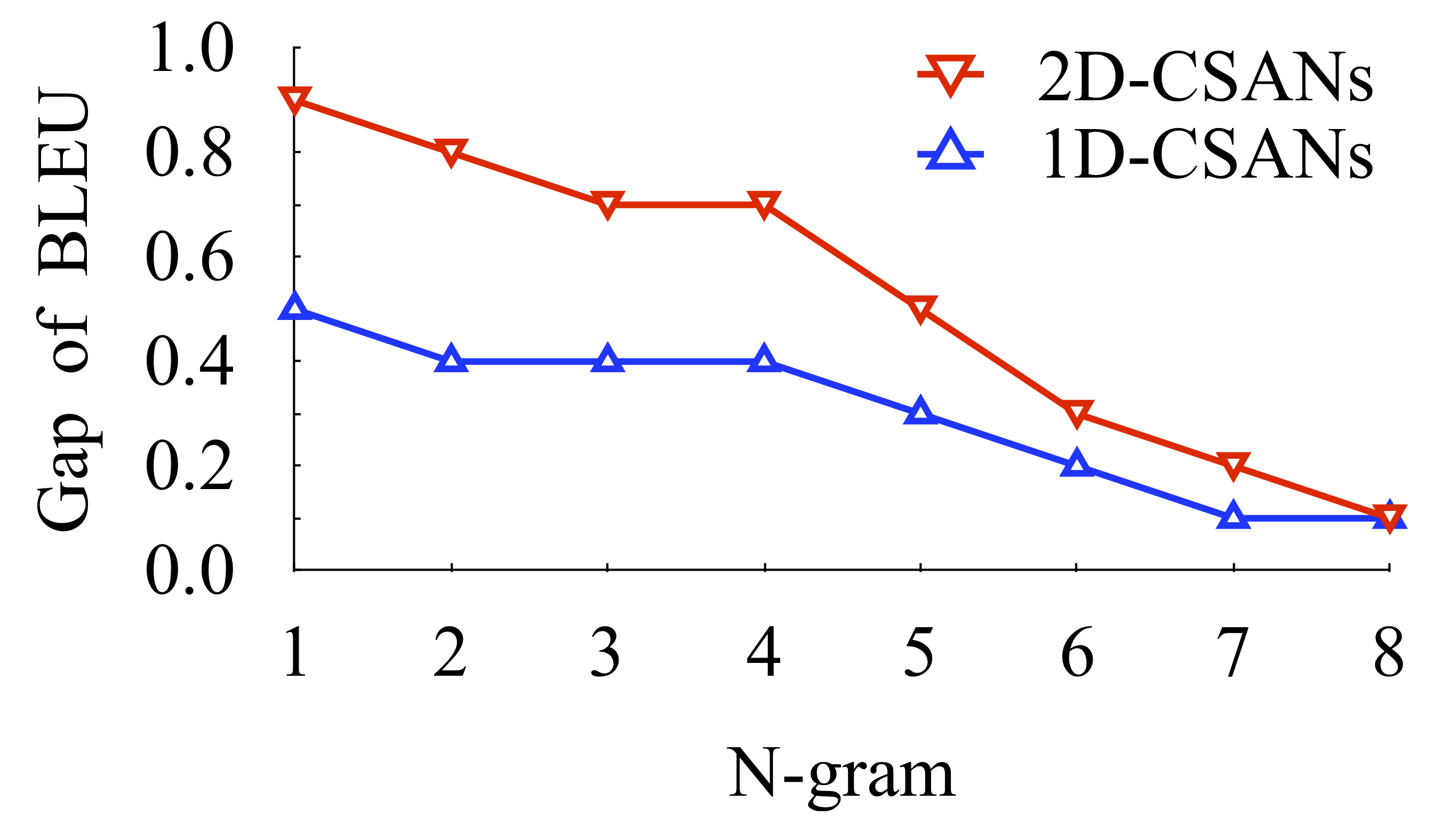}
\caption{Performance improvement on different n-grams. ``Gap of BLEU'' denotes the improvement achieved by the proposed models over the baseline.} 
\label{fig:ana}
\end{center}
\end{figure}

One intuition of our approach is to capture useful phrasal patterns via modeling locality. To evaluate the accuracy of phrase translations, we calculate the improvement of the proposed approaches over multiple granularities of n-grams, as shown in Figure~\ref{fig:ana}.
Both the two model variations consistently outperform the baseline on larger granularities, indicating that modeling locality can raise the ability of self-attention model on capturing the phrasal information. Furthermore, the dependencies among heads can be complementary to the localness modeling, which reveals the necessity of the interaction of features in different subspaces.

\section{Conclusion}
In this paper, we propose a parameter-free convolutional self-attention model to enhance the feature extraction of neighboring elements across multiple heads.
Empirical results of machine 
translation task on a variety of language pairs demonstrate the effectiveness and universality of the proposed methods.
The extensive analyses suggest that: 1) modeling locality is beneficial to SANs; 2) interacting features across multiple heads at attention time can further improve the performance; and 3) to some extent, the dynamic weights are superior to their fixed counterpart (i.e. \textsc{CSan}s vs. \textsc{Cnn}s) on local feature extraction. 

As our approach is not limited to the task of machine translation, it is interesting to validate the proposed model in other sequence modeling tasks, such as reading comprehension, language inference, semantic role labeling, sentiment analysis as well as sentence classification.

\section*{Acknowledgments}

The work was partly supported by the National Natural Science Foundation of China (Grant No. 61672555), the Joint Project of Macao Science and Technology Development Fund and National Natural Science Foundation of China (Grant No. 045/2017/AFJ) and the Multiyear Research Grant from the University of Macau (Grant No. MYRG2017-00087-FST).
We thank the anonymous reviewers for their insightful comments.

\balance
\bibliography{naaclhlt2019}

\begin{thebibliography}{32}
\expandafter\ifx\csname natexlab\endcsname\relax\def\natexlab#1{#1}\fi

\bibitem[{Bahdanau et~al.(2015)Bahdanau, Cho, and Bengio}]{bahdanau2015neural}
Dzmitry Bahdanau, Kyunghyun Cho, and Yoshua Bengio. 2015.
\newblock {Neural} {Machine} {Translation} by {Jointly} {Learning} to {Align}
  and {Translate}.
\newblock In \emph{ICLR}.

\bibitem[{Chen et~al.(2018)Chen, Firat, Bapna, Johnson, Macherey, Foster,
  Jones, Schuster, Shazeer, Parmar, Vaswani, Uszkoreit, Kaiser, Chen, Wu, and
  Hughes}]{Chen:2018:ACL}
Mia~Xu Chen, Orhan Firat, Ankur Bapna, Melvin Johnson, Wolfgang Macherey,
  George Foster, Llion Jones, Mike Schuster, Noam Shazeer, Niki Parmar, Ashish
  Vaswani, Jakob Uszkoreit, Lukasz Kaiser, Zhifeng Chen, Yonghui Wu, and
  Macduff Hughes. 2018.
\newblock The best of both worlds: Combining recent advances in neural machine
  translation.
\newblock In \emph{ACL}.

\bibitem[{Dou et~al.(2018)Dou, Tu, Wang, Shi, and Zhang}]{Dou:2018:EMNLP}
Ziyi Dou, Zhaopeng Tu, Xing Wang, Shuming Shi, and Tong Zhang. 2018.
\newblock {Exploiting Deep Representations for Neural Machine Translation}.
\newblock In \emph{EMNLP}.

\bibitem[{Dou et~al.(2019)Dou, Tu, Wang, Wang, Shi, and Zhang}]{Dou:2019:AAAI}
Ziyi Dou, Zhaopeng Tu, Xing Wang, Longyue Wang, Shuming Shi, and Tong Zhang.
  2019.
\newblock {Dynamic Layer Aggregation for Neural Machine Translation}.
\newblock In \emph{AAAI}.

\bibitem[{Fukui et~al.(2016)Fukui, Park, Yang, Rohrbach, Darrell, and
  Rohrbach}]{fukui2016multimodal}
Akira Fukui, Dong~Huk Park, Daylen Yang, Anna Rohrbach, Trevor Darrell, and
  Marcus Rohrbach. 2016.
\newblock {Multimodal Compact Bilinear Pooling for Visual Question Answering
  and Visual Grounding}.
\newblock In \emph{EMNLP}.

\bibitem[{Guo et~al.(2019)Guo, Zhang, and Liu}]{guo2019gaussian}
Maosheng Guo, Yu~Zhang, and Ting Liu. 2019.
\newblock {Gaussian Transformer: A Lightweight Approach for Natural Language
  Inference}.
\newblock In \emph{AAAI}.

\bibitem[{Hao et~al.(2019)Hao, Wang, Yang, Wang, Zhang, and
  Tu}]{Hao:2019:NAACL}
Jie Hao, Xing Wang, Baosong Yang, Longyue Wang, Jinfeng Zhang, and Zhaopeng Tu.
  2019.
\newblock {Modeling Recurrence for Transformer}.
\newblock In \emph{NAACL}.

\bibitem[{Kim(2014)}]{kim2014convolutional}
Yoon Kim. 2014.
\newblock {Convolutional Neural Networks for Sentence Classification}.
\newblock In \emph{EMNLP}.

\bibitem[{Koehn(2004)}]{Koehn2004Statistical}
Philipp Koehn. 2004.
\newblock {Statistical Significance Tests for Machine Translation Evaluation}.
\newblock In \emph{EMNLP}.

\bibitem[{Li et~al.(2018)Li, Tu, Yang, Lyu, and Zhang}]{Li:2018:EMNLP}
Jian Li, Zhaopeng Tu, Baosong Yang, Michael~R. Lyu, and Tong Zhang. 2018.
\newblock {Multi-Head Attention with Disagreement Regularization}.
\newblock In \emph{EMNLP}.

\bibitem[{Li et~al.(2019)Li, Yang, Dou, Wang, Lyu, and Tu}]{Li:2019:NAACL}
Jian Li, Baosong Yang, Zi-Yi Dou, Xing Wang, Michael~R. Lyu, and Zhaopeng Tu.
  2019.
\newblock {Information Aggregation for Multi-Head Attention with
  Routing-by-Agreement}.
\newblock In \emph{NAACL}.

\bibitem[{Lin et~al.(2014)Lin, Chen, and Yan}]{lin2013network}
Min Lin, Qiang Chen, and Shuicheng Yan. 2014.
\newblock {Network in Network}.
\newblock In \emph{ICLR}.

\bibitem[{Lin et~al.(2017)Lin, Feng, Santos, Yu, Xiang, Zhou, and
  Bengio}]{lin2017structured}
Zhouhan Lin, Minwei Feng, Cicero Nogueira~dos Santos, Mo~Yu, Bing Xiang, Bowen
  Zhou, and Yoshua Bengio. 2017.
\newblock {A Structured Self-Aattentive Sentence Embedding}.
\newblock In \emph{ICLR}.

\bibitem[{Luong et~al.(2015)Luong, Pham, and Manning}]{luong2015effective}
Thang Luong, Hieu Pham, and Christopher~D. Manning. 2015.
\newblock {Effective Approaches to Attention-based Neural Machine Translation}.
\newblock In \emph{EMNLP}.

\bibitem[{Morishita et~al.(2017)Morishita, Suzuki, and
  Nagata}]{morishita2017ntt}
Makoto Morishita, Jun Suzuki, and Masaaki Nagata. 2017.
\newblock {NTT Neural Machine Translation Systems at WAT 2017}.
\newblock In \emph{WAT}.

\bibitem[{Ngiam et~al.(2011)Ngiam, Khosla, Kim, Nam, Lee, and
  Ng}]{ngiam2011multimodal}
Jiquan Ngiam, Aditya Khosla, Mingyu Kim, Juhan Nam, Honglak Lee, and Andrew~Y
  Ng. 2011.
\newblock {Multimodal Deep Learning}.
\newblock In \emph{ICML}.

\bibitem[{Parikh et~al.(2016)Parikh, T{\"a}ckstr{\"o}m, Das, and
  Uszkoreit}]{parikh2016decomposable}
Ankur Parikh, Oscar T{\"a}ckstr{\"o}m, Dipanjan Das, and Jakob Uszkoreit. 2016.
\newblock {A Decomposable Attention Model for Natural Language Inference}.
\newblock In \emph{EMNLP}.

\bibitem[{Peters et~al.(2018)Peters, Neumann, Iyyer, Gardner, Clark, Lee, and
  Zettlemoyer}]{Peters:2018:NAACL}
Matthew~E Peters, Mark Neumann, Mohit Iyyer, Matt Gardner, Christopher Clark,
  Kenton Lee, and Luke Zettlemoyer. 2018.
\newblock {Deep Contextualized Word Representations}.
\newblock In \emph{NAACL}.

\bibitem[{Raganato and Tiedemann(2018)}]{Raganato:2018:EMNLPWorkshop}
Alessandro Raganato and J{\"o}rg Tiedemann. 2018.
\newblock {An Analysis of Encoder Representations in Transformer-Based Machine
  Translation}.
\newblock In \emph{EMNLP Workshop BlackboxNLP: Analyzing and Interpreting
  Neural Networks for NLP}.

\bibitem[{Sennrich et~al.(2016)Sennrich, Haddow, and
  Birch}]{sennrich2015neural}
Rico Sennrich, Barry Haddow, and Alexandra Birch. 2016.
\newblock {Neural Machine Translation of Rare Words with Subword Units}.
\newblock In \emph{ACL}.

\bibitem[{Shaw et~al.(2018)Shaw, Uszkoreit, and Vaswani}]{shaw2018self}
Peter Shaw, Jakob Uszkoreit, and Ashish Vaswani. 2018.
\newblock {Self-Attention with Relative Position Representations}.
\newblock In \emph{NAACL}.

\bibitem[{Shen et~al.(2018{\natexlab{a}})Shen, Zhou, Long, Jiang, Pan, and
  Zhang}]{Shen:2018:AAAI}
Tao Shen, Tianyi Zhou, Guodong Long, Jing Jiang, Shirui Pan, and Chengqi Zhang.
  2018{\natexlab{a}}.
\newblock {DiSAN: Directional Self-Attention Network for RNN/CNN-Free Language
  Understanding}.
\newblock In \emph{AAAI}.

\bibitem[{Shen et~al.(2018{\natexlab{b}})Shen, Zhou, Long, Jiang, and
  Zhang}]{Shen:2018:ICLR}
Tao Shen, Tianyi Zhou, Guodong Long, Jing Jiang, and Chengqi Zhang.
  2018{\natexlab{b}}.
\newblock {Bi-Directional Block Self-Attention for Fast and Memory-Efficient
  Sequence Modeling}.
\newblock In \emph{ICLR}.

\bibitem[{Sperber et~al.(2018)Sperber, Niehues, Neubig, St{\"u}ker, and
  Waibel}]{sperber2018self}
Matthias Sperber, Jan Niehues, Graham Neubig, Sebastian St{\"u}ker, and Alex
  Waibel. 2018.
\newblock {Self-Attentional Acoustic Models}.
\newblock \emph{Interspeech}.

\bibitem[{Strubell et~al.(2018)Strubell, Verga, Andor, Weiss, and
  McCallum}]{Strubell:2018:EMNLP}
Emma Strubell, Patrick Verga, Daniel Andor, David Weiss, and Andrew McCallum.
  2018.
\newblock {Linguistically-Informed Self-Attention for Semantic Role Labeling}.
\newblock In \emph{EMNLP}.

\bibitem[{Vaswani et~al.(2017)Vaswani, Shazeer, Parmar, Uszkoreit, Jones,
  Gomez, Kaiser, and Polosukhin}]{Vaswani:2017:NIPS}
Ashish Vaswani, Noam Shazeer, Niki Parmar, Jakob Uszkoreit, Llion Jones,
  Aidan~N Gomez, {\L}ukasz Kaiser, and Illia Polosukhin. 2017.
\newblock {Attention is All You Need}.
\newblock In \emph{NIPS}.

\bibitem[{Wu et~al.(2018)Wu, Wang, Liu, and Ma}]{wu2018phrase}
Wei Wu, Houfeng Wang, Tianyu Liu, and Shuming Ma. 2018.
\newblock {Phrase-level Self-Attention Networks for Universal Sentence
  Encoding}.
\newblock In \emph{EMNLP}.

\bibitem[{Wu and He(2018)}]{wu2018group}
Yuxin Wu and Kaiming He. 2018.
\newblock {Group Normalization}.
\newblock \emph{arXiv:1803.08494}.

\bibitem[{Yang et~al.(2019)Yang, Li, Wong, Chao, Wang, and Tu}]{Yang:2019:AAAI}
Baosong Yang, Jian Li, Derek~F. Wong, Lidia~S. Chao, Xing Wang, and Zhaopeng
  Tu. 2019.
\newblock {Context-Aware Self-Attention Networks}.
\newblock In \emph{AAAI}.

\bibitem[{Yang et~al.(2018)Yang, Tu, Wong, Meng, Chao, and
  Zhang}]{Yang:2018:EMNLP}
Baosong Yang, Zhaopeng Tu, Derek~F. Wong, Fandong Meng, Lidia~S. Chao, and Tong
  Zhang. 2018.
\newblock {Modeling Localness for Self-Attention Networks}.
\newblock In \emph{EMNLP}.

\bibitem[{Yang et~al.(2017)Yang, Wong, Xiao, Chao, and Zhu}]{yang2017towards}
Baosong Yang, Derek~F Wong, Tong Xiao, Lidia~S Chao, and Jingbo Zhu. 2017.
\newblock {Towards Bidirectional Hierarchical Representations for
  Attention-based Neural Machine Translation}.
\newblock In \emph{EMNLP}.

\bibitem[{Yu et~al.(2018)Yu, Dohan, Luong, Zhao, Chen, Norouzi, and
  Le}]{Yu:2018:ICLR}
Adams~Wei Yu, David Dohan, Minh-Thang Luong, Rui Zhao, Kai Chen, Mohammad
  Norouzi, and Quoc~V Le. 2018.
\newblock {QANet: Combining Local Convolution with Global Self-attention for
  Reading Comprehension}.
\newblock In \emph{ICLR}.

\end{thebibliography}
\bibliographystyle{acl_natbib}
\end{document}